\documentclass[letterpaper]{article} % DO NOT CHANGE THIS
\usepackage[submission]{aaai2026}  % DO NOT CHANGE THIS
\usepackage{times}  % DO NOT CHANGE THIS
\usepackage{helvet}  % DO NOT CHANGE THIS
\usepackage{courier}  % DO NOT CHANGE THIS
\usepackage[hyphens]{url}  % DO NOT CHANGE THIS
\usepackage{graphicx} % DO NOT CHANGE THIS
\usepackage{natbib}  % DO NOT CHANGE THIS AND DO NOT ADD ANY OPTIONS TO IT
\usepackage{caption} % DO NOT CHANGE THIS AND DO NOT ADD ANY OPTIONS TO IT
\usepackage{algorithm}
\usepackage{algorithmic}

\usepackage{amsfonts} % wjy
\usepackage{amsmath} % wjy
\usepackage{multirow} % wjy
\usepackage{multicol} % wjy
\usepackage{pifont} % wjy
\usepackage{subcaption} % wjy

\usepackage{newfloat}
\usepackage{listings}
\DeclareCaptionStyle{ruled}{labelfont=normalfont,labelsep=colon,strut=off} % DO NOT CHANGE THIS
\floatstyle{ruled}
\newfloat{listing}{tb}{lst}{}
\floatname{listing}{Listing}
\title{Pure-Pass: Fine-Grained, Adaptive Masking for\\ Dynamic Token-Mixing Routing in Lightweight Image Super-Resolution}
\author{
Junyu Wu,
Jie Liu\thanks{Jie Liu is the corresponding author
(liujie@nju.edu.cn).},
Jie Tang,
Gangshan Wu\\
State Key Laboratory for Novel Software Technology\\
Nanjing University\\
221870052@smail.nju.edu.cn,
\{liujie, tangjie, gswu\}@nju.edu.cn\\
\texttt{https://github.com/idiotgoose/PurePass}
}
\affiliations{
    \textsuperscript{\rm 1}Association for the Advancement of Artificial Intelligence\\
    1101 Pennsylvania Ave, NW Suite 300\\
    Washington, DC 20004 USA\\
    proceedings-questions@aaai.org
}

\usepackage{bibentry}
\begin{document}

\maketitle

\begin{abstract}
Image Super-Resolution (SR) aims to reconstruct high-resolution images from low-resolution counterparts, but the computational complexity of deep learning-based methods often hinders practical deployment. CAMixer is the pioneering work to integrate the advantages of existing lightweight SR methods and proposes a content-aware mixer to route token mixers of varied complexities according to the difficulty of content recovery. However, several limitations remain, such as poor adaptability, coarse-grained masking and spatial inflexibility, among others. We propose Pure-Pass (PP), a pixel-level masking mechanism that identifies pure pixels and exempts them from expensive computations. PP utilizes fixed color center points to classify pixels into distinct categories, enabling fine-grained, spatially flexible masking while maintaining adaptive flexibility. Integrated into the state-of-the-art ATD-light model, PP-ATD-light achieves superior SR performance with minimal overhead, outperforming CAMixer-ATD-light in reconstruction quality and parameter efficiency when saving a similar amount of computation.
\end{abstract}

% Uncomment the following to link to your code, datasets, an extended version or similar.
% You must keep this block between (not within) the abstract and the main body of the paper.
% \begin{links}
%     \link{Code}{https://aaai.org/example/code}
%     \link{Datasets}{https://aaai.org/example/datasets}
%     \link{Extended version}{https://aaai.org/example/extended-version}
% \end{links}

\section{Introduction}
Image Super-Resolution (SR) represents a fundamental challenge in computer vision, aiming to reconstruct High-Resolution (HR) images with enhanced structural detail and textural fidelity from their Low-Resolution (LR) counterparts. Recent advances in deep learning have significantly improved SR performance \cite{zhang2024transcending, hsu2024drct, ray2024cfat, chu2024hmanet, li2024hierarchical}; however, these gains are often accompanied by substantial increases in computational complexity and resource demands, hindering their practical deployment in real-world applications. To address this limitation, considerable research efforts have been devoted to developing computationally efficient solutions, including network pruning \cite{shi2023memory, zhang2021learning, yu2023dipnet}, quantization \cite{qin2023quantsr, tu2023toward, liu20242dquant, lee2024refqsr}, knowledge distillation \cite{jiang2024mtkd, xie2023large}, and the design of lightweight architectures \cite{wang2023omni, li2023efficient, choi2023n, zhang2024hit, huang2025feature}. While these techniques have demonstrated success in accelerating inference for resource-constrained platforms, they predominantly employ static network structures that process all input regions uniformly, failing to account for the intrinsic variability in image content that could enable more efficient discriminative processing.

In recent years, accelerating frameworks \cite{chen2022arm, kong2021classsr, nguyen2025enaf, wang2024psar, yu2021path, yu2024rethinking, hu2022restore} have emerged as a prominent research direction in efficient SR. This strategy stems from the observation that different image regions have different restoration difficulties and can be processed by networks with different capacities. The core methodology involves decomposing images into fixed sub-images and employing specialized SR networks tailored to the restoration requirements of each sub-image. ARM \cite{chen2022arm} and WBSR \cite{yu2024rethinking} further advanced the strategy by introducing a supernet architecture that dynamically allocates subnet models without introducing additional parameters. However, a critical limitation of this strategy lies in its constrained receptive fields. The decomposition of images into sub-images inherently restricts the receptive field of the network, which adversely affects restoration performance. Moreover, poor classification and inflexible partition also undermine its performance, as shown in \cite{wang2024camixersr}.

CAMixer \cite{wang2024camixersr} is the pioneering work to integrate the advantages of above strategies. Based on the derived observation that distinct feature regions demand varying levels of token-mixer complexities (e.g., simple convolutions for smooth areas and complex self-attention for textures), it proposes a content-aware mixer to route token mixers with different complexities according to the content. At the implementation level, CAMixer proposes a computation-saving approach for Window Self-Attention: it first utilizes a predictor to generate a window-level mask; when the predicted mask identifies a window as requiring simple super-resolution, the computationally expensive Window Self-Attention operation is replaced with a more efficient convolutional alternative to reduce computational overhead. This adaptive routing mechanism achieves substantial computational savings with marginal degradation in reconstruction quality, striking an improved balance between efficiency and performance in Transformer-based super-resolution.

Despite its innovative concept, the effectiveness of CAMixer's specific implementation remains underexplored, especially when applied to more advanced, SOTA lightweight architectures. To investigate this, we integrated CAMixer into ATD-light (Zhang et al. 2024), a powerful model featuring three complementary parallel attention mechanisms: (1) Adaptive Token Dictionary Cross-Attention (ATD-CA), (2) Adaptive Category-based Multi-head Self-Attention (AC-MSA), and (3) Shifted Window-based Multi-head Self-Attention (SW-MSA) modules. This hybrid design pursued two goals: first, to reduce ATD-light’s computational overhead while maintaining its reconstruction quality, and second, to evaluate CAMixer’s potential in more complex frameworks. Specifically, in coordination with CAMixer's implementation idea, we utilize CAMixer's window-level mask to save calculation overload for the SW-MSA module. However, this integration proved counterproductive. As shown in Tables 1 and 2, the resulting CAMixer-ATD-light model suffered a significant performance drop while paradoxically increasing the parameter count. This unexpected failure motivated a critical re-examination of CAMixer's underlying design.

Our analysis reveals that CAMixer's shortcomings stem from five fundamental design flaws: \textbf{(1) Poor Adaptability}: It relies on a fixed ratio, failing to adapt computation savings to the input image's actual complexity. \textbf{(2) Coarse-grained Masking}: The mask resolution is coupled to the Window Attention's fixed $16\times16$ window size, which is too coarse to accurately distinguish fine-grained textures. \textbf{(3) Spatial Scale Inflexibility of Masking}: Its fixed grid partitioning struggles with textures located at window intersections. \textbf{(4) Incompatibility with SW-MSA}: It forces the disabling of the crucial window-shifting mechanism in modern Transformers. \textbf{(5) High Overhead}: It introduces a non-trivial parametric and computational burden. (These are detailed in the Related Work section).

To mitigate the aforementioned limitations, we propose a novel approach termed Pure-Pass (PP). In contrast to CAMixer’s window-level masking strategy, our method introduces a more fine-grained pixel-level masking mechanism that selectively identifies \textit{pure pixels} within homogeneous regions using a novel method based on fixed color centers. These identified pixels are then exempted from expensive computations in the AC-MSA module of ATD-light. To ensure fidelity, a zero-cost compensation mechanism re-integrates information from the parallel SW-MSA branch for these bypassed pixels.

% Compared with CAMixer, our PP approach offers (1) Adaptive Flexibility, (2) Fine-Grained Masking, (3) Spatially Flexible Masking, (4) Compatibility-Preserving Optimization for SW-MSA, and (5) Negligible Parametric and Computational Overhead.

% Comparative evaluations demonstrate that our PP-ATD-light architecture achieves significantly superior super-resolution performance compared to CAMixer-ATD-light when saving a similar amount of computation. Notably, these performance improvements are attained with a reduced parameter count compared to the CAMixer-based approach. The observed superiority can be attributed to the overcome of limitations inherent in CAMixer.

The design of Pure-Pass directly addresses the identified flaws. It offers: \textbf{(1) Adaptive Flexibility} by dynamically determining the amount of computation to save based on image content. \textbf{(2) Fine-Grained Masking} by decoupled window processing from Window Attention for adaptive texture analysis. \textbf{(3) Spatially Flexible Masking} through a cross-shift fusion strategy. \textbf{(4) Compatibility-Preserving Optimization for SW-MSA} by targeting a different module (AC-MSA) with pixel-level masking. \textbf{(5) Negligible Overhead} in parameters and computation. (These are detailed in the Method section). Consequently, our PP-ATD-light significantly outperforms CAMixer-ATD-light in both reconstruction quality and parameter efficiency, while achieving comparable or even greater computational savings.

\section{Related Work}
\subsection{Accelerating Framework for SR}
As the architectural complexity of super-resolution models continues to escalate to achieve superior restoration quality, their practical deployment faces increasing challenges. Accelerating framework \cite{kong2021classsr, chen2022arm, yu2021path, yu2024rethinking, wang2024psar, nguyen2025enaf, hu2022restore} addresses this problem by adopting content-aware modules to dynamically send sub-images (patches) to sub-networks with different complexities to accelerate model inference. PathRestore \cite{yu2021path} introduces a pathfinder to implement a multi-path CNN, selecting feature paths to adapt FLOPs according to context.  ClassSR \cite{kong2021classsr} develops a three-class classifier that directs input patches to either complex, medium, or simple processing networks, which saves 31\% calculations for SRResNet \cite{ledig2017photo} on 2K datasets. ARM \cite{chen2022arm} utilizes supernet to share parameters among subnets and it further builds an Edge-to-PSNR lookup table that correlates patch-level edge complexity with subnet performance metrics, enabling intelligent and efficient subnet selection. WBSR \cite{yu2024rethinking} employs gradient projection maps as an alternative to conventional edge detection methods to facilitate accurate reconstruction. ENAF \cite{nguyen2025enaf} further improves the framework with multiple early exits and a compact PSNR estimator to make a better computational-performance trade-off.

\subsection{Dynamic Token-Mixing Routing}
Dynamic token-mixing routing stems from the observation that different feature regions inherently require varying degrees of token-mixer complexity. In the context of super-resolution, CAMixer \cite{wang2024camixersr} pioneers this approach by adaptively routing tokens to either complex self-attention mechanisms for intricate regions or simpler convolutional operations for homogeneous areas, based on local content characteristics. Cat-AIR \cite{jiang2025cat} further transfers the idea of dynamic token-mixing routing to all-in-one image restoration, which incorporates both content and task awareness, enabling efficient processing across multiple restoration tasks.

\begin{figure}[t]
\centering
\includegraphics[width=0.5\columnwidth]{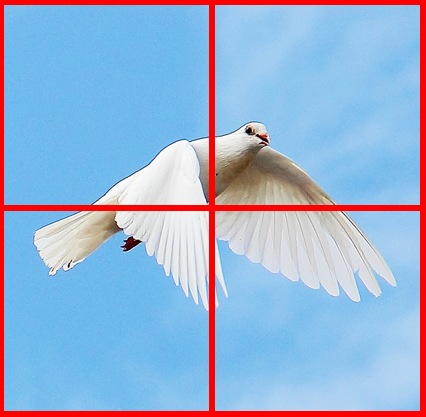} 
\caption{An example of localized regions containing complex textures positioned at the intersection points of four adjacent windows within a cross-shaped configuration, while being surrounded by homogeneous, textureless areas.}
\label{fig_bird_sky_grid}
\end{figure}

Despite CAMixer's pioneering approach, we identify several key limitations in its design that motivate our work:
\begin{enumerate}
    \item \textbf{Poor Adaptability and Flexibility}: CAMixer relies on a predefined "ratio" hyperparameter, which denotes the proportion of windows with higher computational complexity. This results in its inability to dynamically determine the optimal proportion of regions requiring high-complexity computation based on the characteristics and features of the input image. Consequently, two notable limitations emerge: (1) For images dominated by extensive homogeneous regions, the model fails to maximize potential computational savings, and (2) For images containing predominantly intricate textures, excessive computational reduction may occur, thereby compromising restoration quality.
    \item \textbf{Coarse-grained Masking}: In CAMixer, the window size for generating window-level masks is inherently constrained by that of the Window Self-Attention mechanism. However, existing approaches typically employ a fixed window size of $16 \times 16$ for Window Self-Attention due to the trade-off between computational efficiency and super-resolution performance. This architectural constraint consequently compels CAMixer to adopt the same $16 \times 16$ window configuration for mask generation. Such a design limitation results in a suboptimal, coarse-grained discrimination capability for region-wise super-resolution complexity assessment.
    \item \textbf{Spatial Scale Inflexibility of Masking}: Not only the window size of window-level mask in CAMixer is fixed, but also the spatial positions of the windows are rigid, preventing dynamic adjustments. This architectural constraint becomes particularly problematic when processing heterogeneous image regions—for example, when localized regions containing complex textures happen to be positioned at the intersection points of four adjacent windows within a cross-shaped configuration, while being surrounded by homogeneous, textureless areas, as shown in Figure \ref{fig_bird_sky_grid}. In such cases, the model cannot dynamically adjust window boundaries to focus computational resources on these critical regions, leading to either insufficient attention to complex textures or wasteful computation on smooth areas. While the flow-warped key tokens in W-MSA computation provide partial mitigation, this remains a superficial solution as the fundamental issue of spatially fixed window partitioning persists.
    % \item \textbf{Limited Accuracy and Interpretability of Masking}: The masks generated by CAMixer lack sufficient precision, potentially misidentifying and erroneously masking regions with complex textures that actually require high-complexity computations, as illustrated in Figure \ref{}. Furthermore, CAMixer employs learnable neural networks to predict masks, resulting in poor interpretability.
    \item \textbf{Incompatibility with Shifted Window-based Multi-head Self-Attention (SW-MSA)}: As a computation-efficient approach designed for standard W-MSA, CAMixer generates window-level masks. However, shifted window operations, which typically displace windows by half their length, disrupt the original window alignment, rendering the precomputed masks invalid. Consequently, we are compelled to disable the window-shifting mechanism in SW-MSA, thus sacrificing performance.
    \item \textbf{Additional Parametric and Computational Costs}: The integration of CAMixer for ATD-light introduces additional model complexity, increasing the total parameter count by approximately 10\%. Moreover, the CAMixer module itself accounts for roughly 4\% of the original model’s computational load.   
\end{enumerate}

\section{Method}

\subsection{Pure-Pass: Pixel-level Mask Generation }
\begin{figure*}[t]
    \centering
    \includegraphics[width=0.9\textwidth]{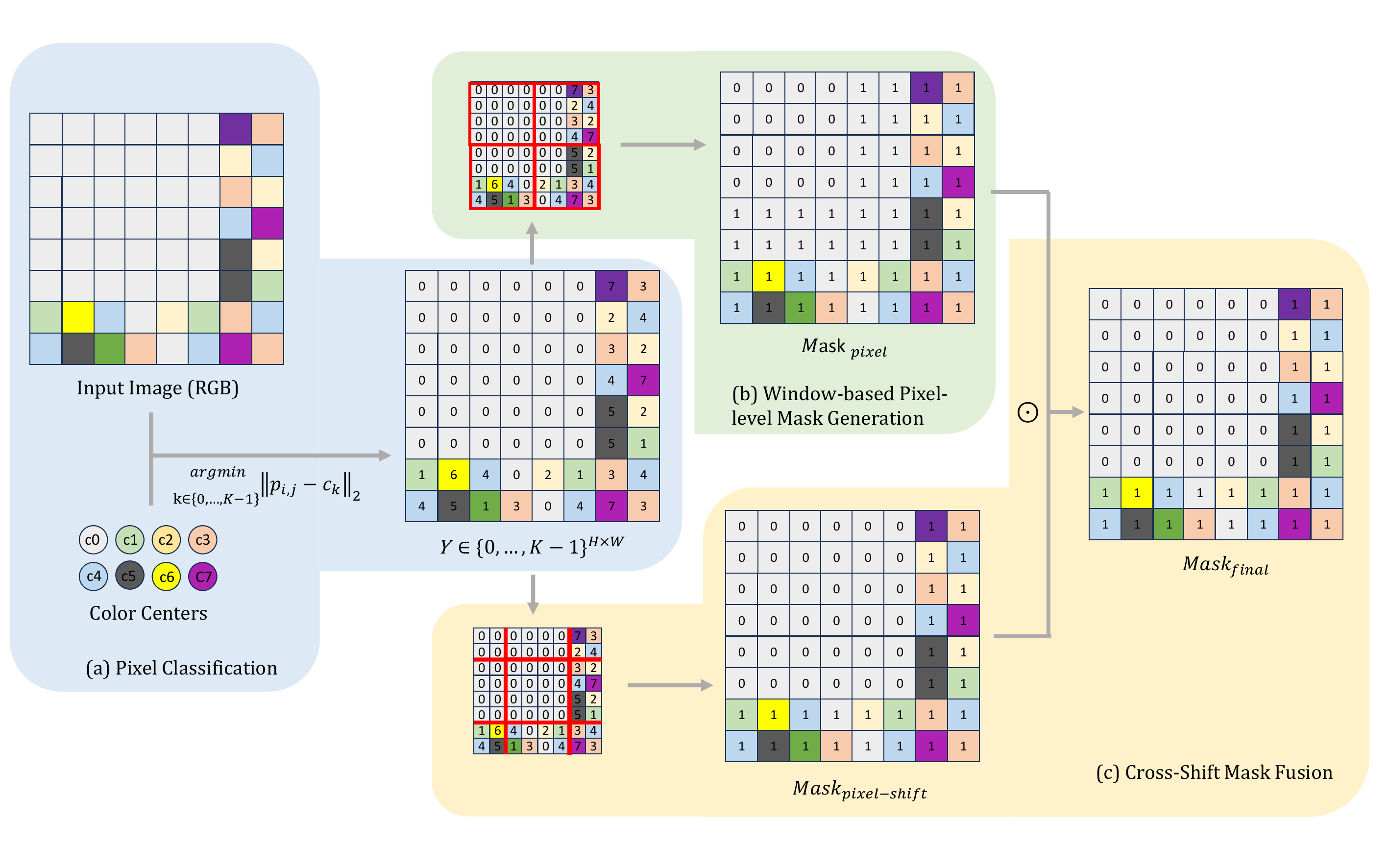}
    \caption{Pure-Pass mask generation pipeline: (a) Pixel Classification, (b) Window-based Pixel-level Mask Generation, (c) Cross-Shift Mask Fusion}
    \label{fig:pp_framework}
\end{figure*}

We introduce a novel \textbf{Pure-Pass (PP)} method to efficiently identify texture-deficient regions in input images. Our method generates pixel-level masks through a three-stage process, as illustrated in Figure \ref{fig:pp_framework}.

\subsubsection{Pixel Classification}
We first define a set of $K$ fixed color centers $\mathbf{C} = \{\mathbf{c}_1, \mathbf{c}_2, ..., \mathbf{c}_K\}$, where each center $\mathbf{c}_k \in \mathbb{R}^3$ represents an RGB color value. These centers are uniformly distributed in HSV color space for better perceptual uniformity and then converted to RGB space:

\begin{equation}
    \mathbf{c}_k = \text{HSVtoRGB}(h_k, s_k, v_k)
\end{equation}
where:
\begin{itemize}
    \item $h_k = \frac{k}{K}$ is the hue value for the $k$-th center (uniformly spaced in $[0,1]$)
    \item $s_k = 0.9$ and $v_k = 0.9$ are fixed saturation and value parameters
\end{itemize}

Given an input image $\mathbf{I} \in \mathbb{R}^{H \times W \times C_0}$, where $H$ denotes the height, $W$ the width, and $C_0=3$ represents the RGB color channels, we define $\mathbf{p}_{i,j} \in [0,1]^3$ as the normalized RGB values of the pixel at spatial location $(i,j)$. 

For each pixel, we assign it to the nearest color center $\mathbf{c}_k$ based on Euclidean distance:

\begin{equation}
    y_{i,j} = \operatorname*{argmin}_{k \in \{0,\ldots,K-1\}} \|\mathbf{p}_{i,j} - \mathbf{c}_k\|_2,
\end{equation}
where:
\begin{itemize}
    \item $y_{i,j}$ is the assigned color category index for pixel $(i,j)$ (ranging from 0 to $K-1$)
    \item $\|\cdot\|_2$ represents Euclidean distance
\end{itemize}

The resulting assignment forms a label map $\mathbf{Y} \in \{0,\ldots,K-1\}^{H \times W}$ that serves as the foundation for subsequent processing.

\subsubsection{Window-based Pixel-level Mask Generation}
Building upon the pixel classification results obtained in the previous stage, we generate pixel-level masks using a window-based approach to identify texture-deficient regions. For an input image of dimensions $H \times W$, we partition it into non-overlapping square windows of size $S \times S$.

Each window $\mathbf{W}_{m,n}$, positioned at grid coordinates $(m,n)$, is defined as:

\begin{equation}
    \mathbf{W}_{m,n} = \mathbf{Y}[mS:(m+1)S, nS:(n+1)S].
\end{equation}

We then analyze the color category distribution within each window to determine its texture characteristics. A window is classified as texture-deficient (pure color) if and only if all its pixels share the same color category:

\begin{equation}
    \text{Mask}_{\text{window}}(m,n) = 
    \begin{cases}
    0 & \text{if } \forall (i,j) \in \mathbf{W}_{m,n}, y_{i,j} = y_{mS,nS}, \\
    1 & \text{otherwise},
    \end{cases}
\end{equation}

where:
\begin{itemize}
    \item $\text{Mask}_{\text{window}}(m,n)$ is the window-level mask value (0 for pure color, 1 otherwise)
    \item $y_{mS,nS}$ is the color category of the top-left pixel in the window
\end{itemize}

This window-level mask is then expanded to pixel-level resolution by assigning the same value to all pixels within each window:

\begin{equation}
    \text{Mask}_{\text{pixel}}(i,j) = \text{Mask}_{\text{window}}(\lfloor i/S \rfloor, \lfloor j/S \rfloor).
\end{equation}

\subsubsection{Cross-Shift Mask Fusion}
To enhance the spatial flexibility of our PP method, we introduce a Cross-Shift Mask Fusion strategy. This approach combines masks generated from shifted windows to avoid the dilemma shown in Figure \ref{fig_bird_sky_grid}.

Given the initial pixel-level mask $\text{Mask}_{\text{pixel}}$ generated with window size $S \times S$, we compute a shifted version $\text{Mask}_{\text{pixel-shift}}$ by first shifting the input image by $\lfloor S/2 \rfloor$ pixels in both height and width directions:

\begin{equation}
    \mathbf{Y}_{\text{shift}} = \text{Shift}(\mathbf{Y}, (\delta, \delta)), \quad \delta = \lfloor S/2 \rfloor,
\end{equation}

where:
\begin{itemize}
    \item $\text{Shift}(\cdot)$ denotes the cyclic shift operation
    \item $\delta$ represents the shift size (typically half the window size)
\end{itemize}

The shifted mask $\text{Mask}_{\text{pixel-shift}}$ is then computed using the same window-based procedure on $\mathbf{Y}_{\text{shift}}$. 

The final fused mask is obtained through element-wise multiplication (logical AND) of the base and shifted masks:

\begin{equation}
    \text{Mask}_{\text{final}} = \text{Mask}_{\text{pixel}} \odot \text{Mask}_{\text{pixel-shift}},
\end{equation}

where:
\begin{itemize}
    \item $\odot$ denotes element-wise multiplication
    \item $\text{Mask}_{\text{final}}$ preserves pixels identified as pure-color in either configurations
\end{itemize}

\subsection{PP-ATD-light: Pure-Pass Accelerated ATD-light}
The ATD-light architecture \cite{zhang2024transcending} constitutes a cutting-edge lightweight super-resolution framework that employs triple parallel attention mechanisms for superior detail reconstruction. Our proposed method enhances this framework through strategic computation bypassing for AC-MSA in texture-deficient regions, as identified by the Pure-Pass algorithm, while maintaining reconstruction fidelity via an innovative compensation mechanism.

\subsubsection{Original AC-MSA Procedure} 
The conventional AC-MSA operation is formally expressed as:
\begin{align}
    \{\phi^j\} &= \text{Categorize}\{X, A\}, \\
    \hat{\phi^j} &= \text{MSA}(\phi^jW^Q, \phi^jW^K, \phi^jW^V), \\
    X_{out} &= \text{UnCategorize}(\{\hat{\phi^j}\}),
\end{align}
where:
\begin{itemize}
    \item $A \in \mathbb{R}^{HW \times M}$ denotes the similarity map derived from ATD-CA
    \item $X \in \mathbb{R}^{HW \times C}$ represents all image tokens
    \item $\phi^j$ corresponds to the j-th sub-category (j = 0, 1, 2...)
\end{itemize}

\subsubsection{Pure-Pass Optimized AC-MSA}
Upon detection of \textit{pure pixels} through our Pure-Pass algorithm, we eliminate redundant computations for these regions within the AC-MSA module while preserving essential processing for textured areas.

The optimization process begins with extracting indices of \textit{non-pure pixels}:
\begin{equation}
    I_{hard} = \text{mask2index}(\text{Mask}_{\text{final}}).
\end{equation}

Subsequently, we perform selective processing on only the relevant pixels:
\begin{align}
    X_{hard} &= X[I_{hard}], \\
    A_{hard} &= A[I_{hard}], \\
    \{\phi_{hard}^j\} &= \text{Categorize}\{X_{hard}, A_{hard}\}, \\
    \hat{\phi_{hard}^j} &= \text{MSA}(\phi_{hard}^jW^Q, \phi_{hard}^jW^K, \phi_{hard}^jW^V), \\
    X_{out-hard} &= \text{UnCategorize}(\{\hat{\phi_{hard}^j}\}).
\end{align}

\subsubsection{Information-Preserving Compensation}
To preserve the complete information volume, we introduce a compensation mechanism that leverages the outputs from SW-MSA without introducing extra calculation:
\begin{align}
    X_{comp} &= X_{SW\text{-}MSA} \odot (1-\text{Mask}_{\text{final}}), \\
    \hat{X_{out}} &= \text{PutTogether}(X_{out\text{-}hard}, X_{comp}),
\end{align}
where $X_{SW\text{-}MSA}$ represents the output from the SW-MSA module. The compensation term $X_{comp}$ contains feature information from \textit{pure pixels}, which is then combined with the processed non-pure features to form the complete output representation.

The observed superiority can be attributed to the overcome of limitations inherent in CAMixer:
\subsection{Discussion: Advantages}
The proposed Pure-Pass method overcomes CAMixer’s limitations through five key advantages:
\begin{enumerate}
    \item \textbf{Adaptive Flexibility}: Our model incorporates an adaptive mechanism that autonomously determines the proportion of regions requiring less intensive computation based on the inherent features and characteristics of the input images.
    \item \textbf{Fine-Grained Masking}: Our masks operate at the pixel level. Although we leverage window-based processing for mask generation, these windows are decoupled and independent from those used in Window Attention. Unlike CAMixer, which is constrained by fixed window sizes, our approach allows for smaller windows to more precisely assess whether a region contains complex textures and requires intensive computation.
    \item \textbf{Spatially Flexible Masking}: Although our mask generation process utilizes window-based analysis, the resulting masks maintain pixel-level granularity. Consequently, we can break fixed spatial constraints by incorporating shifted-window operations, where masks from both original and shifted windows are adaptively combined. As a result, our method achieves superior spatial flexibility in the final mask representation.
    % \item \textbf{Enhanced Mask Accuracy and Interpretability}: Our method outperforms CAMixer in mask accuracy while maintaining superior interpretability through explicit color-space decision rules, ensuring full transparency in the mask generation process.
    \item \textbf{Compatibility-Preserving Optimization for SW-MSA}: Our PP method generates pixel-level masks, enabling computational optimization for non-Window-Attention modules. Within the ATD-light architecture, which incorporates three parallel attention modules, the computational load is primarily dominated by AC-MSA and SW-MSA operations. Rather than compromising the window-shifting mechanism of SW-MSA through simplified complexity reduction, we implement a targeted optimization strategy that exclusively enhances the computational efficiency of AC-MSA. This methodology preserves the complete functionality of SW-MSA while achieving substantial computational savings.
    \item \textbf{Negligible Parametric and Computational Overhead} The incorporation of PP \textit{per se} increases the parameter count by less than 1K and the FLOPs by under 0.001G, which is negligible accounting for ATD-light’s total parameter budget and computational load as shown in Table \ref{tab:tab2}.
\end{enumerate}

\begin{figure*}[htbp]
    \centering
    
    % 第一行
    \begin{subfigure}{0.25\textwidth}
        \includegraphics[width=\linewidth]{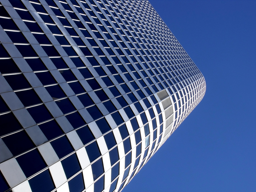}
        \caption{original image}
        \label{fig:mask_compare_1a}
    \end{subfigure}
    \hfill
    \begin{subfigure}{0.25\textwidth}
        \includegraphics[width=\linewidth]{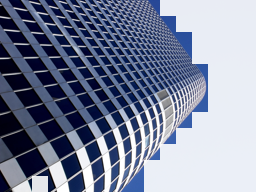}
        \caption{mask by CAMixer}
        \label{fig:mask_compare_1b}
    \end{subfigure}
    \hfill
    \begin{subfigure}{0.25\textwidth}
        \includegraphics[width=\linewidth]
        {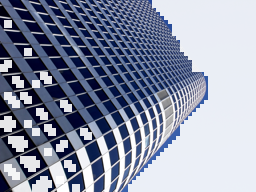}
        \caption{mask by PP}
        \label{fig:mask_compare_1c}
    \end{subfigure}\\
    
    % 第二行
    \begin{subfigure}{0.25\textwidth}
        \includegraphics[width=\linewidth]{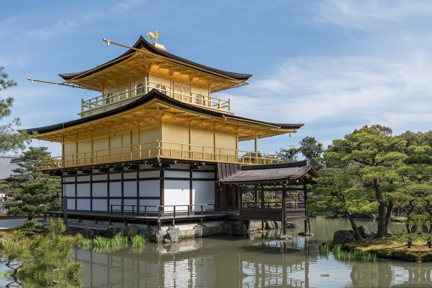}
        \caption{original image}
        \label{fig:mask_compare_2a}
    \end{subfigure}
    \hfill
    \begin{subfigure}{0.25\textwidth}
        \includegraphics[width=\linewidth]{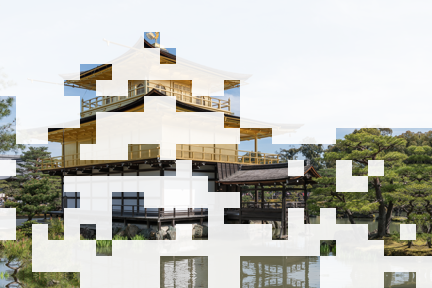}
        \caption{mask by CAMixer}
        \label{fig:mask_compare_2b}
    \end{subfigure}
    \hfill
    \begin{subfigure}{0.25\textwidth}
        \includegraphics[width=\linewidth]{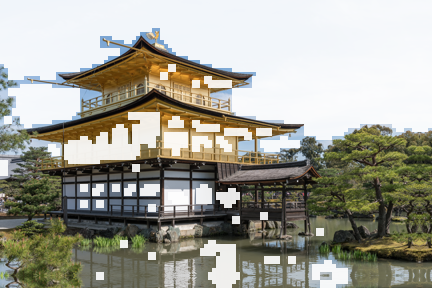}
        \caption{mask by PP}
        \label{fig:mask_compare_2c}
    \end{subfigure}
    
    \caption{Masking Comparison. The white masks in (b), (c), (e), (f) represent areas considered easy for SR, where computational overhead will be saved. }
    \label{fig:mask_compare}
\end{figure*}

\section{Experiment}
\subsection{Experimental Settings}
Our model architecture follows ATD-light \cite{zhang2024transcending}, consisting of 4 ATD blocks. Each block contains six transformer layers with 48 channels. The dictionary size is set to 64 tokens with a reduction rate of 6. For the AC-MSA branch, we use a sub-categories size of 128, while the SW-MSA branch employs a window size of 16. In the Pure-Pass module, we configure a window size of 8, shift size of 4, and maintain 16 fixed color centers.
We train our model on the DIV2K dataset \cite{timofte2017ntire}, and evaluate our model on five standard super-resolution benchmarks: Set5 \cite{bevilacqua2012low}, Set14 \cite{zeyde2010single}, B100 \cite{martin2001database}, Urban100 \cite{huang2015single}, and Manga109 \cite{matsui2017sketch}. Performance is quantified using PSNR and SSIM metrics \cite{wangzhou2004image}, computed on the Y channel after converting images to YCbCr color space. Details of the training and testing procedure can be found in the supplementary material.

\subsection{Masking Comparison with CAMixer}
In the Introduction section, we have highlighted the advantages of the masks generated by our proposed PP method compared to those produced by CAMixer. Here, we present empirical evidence through comparative visualizations to substantiate these claims.

The first row of Figure \ref{fig:mask_compare} demonstrates the superior performance of PP-generated masks. As illustrated in Figure \ref{fig:mask_compare_1b}, the CAMixer-derived mask fails to identify homogeneous regions such as the uniform blue sky areas near the building edges and the black building window regions. In contrast, Figure \ref{fig:mask_compare_1c} shows that mask generated by PP is more fine-grained and spatially flexible, enabling the identification of a greater number of computationally efficient regions for SR and thus saving more computational overhead. 

The second row of Figure \ref{fig:mask_compare} further illustrates the limitations of CAMixer’s fixed masking ratio, which results in unreasonable masking of textured regions such as vegetation and temple eaves (Figure \ref{fig:mask_compare_2b}). Our PP method, however, exhibits dynamic adaptability by selectively masking fewer areas (Figure \ref{fig:mask_compare_2c}), thereby preserving computational resources for regions that truly benefit from SR processing.

\begin{table*}
    \centering
    % Adjust table font size (to a minimum of 9pt)if needed
    \fontsize{9}{11}\selectfont  % Use 9pt if necessary

    % Compress columns if needed
    %\setlength{\tabcolsep}{1mm}  % Reduce default spacing (optional)
    
    \begin{tabular}{c|c|cc|cc|cc|cc|cc}
       \hline
       \multirow{2}{*}{Method}  & \multirow{2}{*}{Scale} & \multicolumn{2}{c|}{Set5} & \multicolumn{2}{c|}{Set14} & \multicolumn{2}{c|}{B100} & \multicolumn{2}{c|}{Urban100} & \multicolumn{2}{c}{Manga109} \\
       && PSNR & SSIM & PSNR & SSIM & PSNR & SSIM & PSNR & SSIM & PSNR & SSIM\\
       \hline
       ATD-light & $\times 2$ & 38.28 & 0.9614 & 34.13 & 0.9222 & 32.39 & 0.9021 & 33.26 & 0.9372 & 39.49 & 0.9790\\
       CAMixer-ATD-light-0.8 & $\times 2$ & 38.23 & 0.9613 & 33.92 & 0.9211 & 32.32 & 0.9013 & 32.94 & 0.9345 & 39.34 & 0.9783\\
       CAMixer-ATD-light-0.5 & $\times 2$ & 38.19 & 0.9612 & 33.87 & 0.9206 & 32.31 & 0.9011 & 32.80 & 0.9335 & 39.29 & 0.9782\\
       \textbf{PP-ATD-light} & $\times 2$ & 38.26 & 0.9615 & 34.17 & 0.9222 & 32.38 & 0.9021 & 33.26 & 0.9375 & 39.50 & 0.9789\\
       \hline
       ATD-light & $\times 3$ & 34.74 & 0.9300 & 30.68 & 0.8484 & 29.31 & 0.8107 & 29.14 & 0.8706 & 34.58 & 0.9504\\
       CAMixer-ATD-light-0.8 & $\times 3$ & 34.60 & 0.9289 & 30.58 & 0.8461 & 29.26 & 0.8090 & 28.82 & 0.8638 & 34.33 & 0.9488\\
       CAMixer-ATD-light-0.5 & $\times 3$ & 34.59 & 0.9288 & 30.57 & 0.8460 & 29.24 & 0.8087 & 28.74 & 0.8628 & 34.28 & 0.9486\\
       \textbf{PP-ATD-light} & $\times 3$ & 34.70 & 0.9296 & 30.64 & 0.8481 & 29.31 & 0.8108 & 29.18 & 0.8709 & 34.52 & 0.9502\\
       \hline
       ATD-light & $\times 4$ & 32.58 & 0.8992 & 28.87 & 0.7884 & 27.78 & 0.7438 & 26.97 & 0.8099 & 31.46 & 0.9198\\
       CAMixer-ATD-light-0.8 & $\times 4$ & 32.45 & 0.8980 & 28.80 & 0.7854 & 27.73 & 0.7414 & 26.71 & 0.8022 & 31.20 & 0.9165\\
       CAMixer-ATD-light-0.5 & $\times 4$ & 32.45 & 0.8978 & 28.78 & 0.7851 & 27.72 &  0.7412 & 26.64 & 0.8006 & 31.14 & 0.9158\\
       \textbf{PP-ATD-light} & $\times 4$ & 32.57 & 0.8991 & 28.91 & 0.7886 & 27.78 & 0.7434 & 26.97 & 0.8102 & 31.40 & 0.9187\\
       \hline
    \end{tabular}
    \caption{Quantitative comparison of super-resolution performance across different datasets and scale factors. The proposed PP-ATD-light maintains competitive performance with the original ATD-light while significantly outperforming both CAMixer variants, particularly on the challenging Urban100 dataset.}
    \label{tab:tab1}
\end{table*}

\begin{table}
    \centering
    \fontsize{9}{11}\selectfont
    \begin{tabular}{|c|c|c|}
        \hline
        Model & Params & FLOPs\\
        \hline
        ATD-light & 769K & 87.15G\\
        \hline
        CAMixer-ATD-light-0.8 & 838K & 83.67G\\
        \hline
        CAMixer-ATD-light-0.5 & 838K & 73.48G\\
        \hline
        \multirow{2}{*}{\textbf{PP-ATD-light}} & \multirow{2}{*}{769K} & 79.30G (average)\\
        && 69.09G (best case)\\
        \hline
    \end{tabular}
    \caption{Model complexity comparison showing parameter counts and computational requirements. PP-ATD-light achieves superior efficiency through its adaptive computation mechanism, maintaining the parameter count of ATD-light while reducing FLOPs by 9\% on average and up to 21\% in optimal cases compared to the original ATD-light.}
    \label{tab:tab2}
\end{table}

\subsection{Performance and Efficiency Comparisons with CAMixer-ATD-light}

Tables \ref{tab:tab1} and \ref{tab:tab2} present comprehensive comparisons between our proposed PP-ATD-light and two baseline models: the original ATD-light and CAMixer-ATD-light variants. In CAMixer-ATD-light-$ratio$, the $ratio$ hyperparameter determines the proportion of windows performing full self-attention computations, with higher values indicating more computationally intensive processing. Notably, CAMixer-ATD-light-0.5 exhibits inferior performance compared to CAMixer-ATD-light-0.8, demonstrating the trade-off between computational savings and model accuracy.

As evidenced in table \ref{tab:tab1}, PP-ATD-light achieves comparable performance to ATD-light across all benchmark datasets, while both CAMixer-ATD-light variants show significant performance degradation. This performance gap is particularly pronounced on the Urban100 dataset, where CAMixer-ATD-light-0.8 and CAMixer-ATD-light-0.5 exhibit PSNR drops of 0.32 dB and 0.46 dB respectively compared to our method on $\times 2$ scale.

The efficiency advantages of our approach are detailed in table \ref{tab:tab2}. PP-ATD-light maintains the parameter efficiency of the original ATD-light (769K parameters), while CAMixer variants require 9\% more parameters (838K).
When it comes to the FLOPs, as PP incorporates an adaptive mechanism that autonomously determines the proportion of regions requiring less intensive computation based on the inherent characteristics of the input images, the percentage of \textit{pure pixels} for PP-ATD-light is averaged on Manga109 x2 dataset when calculating FLOPs. Table \ref{tab:tab2} shows that the average FLOPs of 79.30G for PP-ATD-light is smaller than the FLOPs of 83.67G for CAMixer-ATD-light-0.8. To demonstrate the potential of the adaptive flexibility feature of PP, we also provide the maximal percentage of \textit{pure pixels} of 89.5\% when processing an image from B100. And under this percentage, the FLOPs decrease significantly to 69.09G, outperforming even the more aggressive CAMixer-ATD-light-0.5 (73.48G). Visual examples for qualitative comparisons are provided in the supplementary material.

\begin{table}
    \centering
    \fontsize{9}{11}\selectfont
    \setlength{\tabcolsep}{1mm}
    \begin{tabular}{|c|c|cc|cc|}
        \hline
        Model & Params & \textit{Pure Pixels} & FLOPs & Urban100 & Manga109\\
        \hline
        ATD-light & 769K & 0\% & 87.15G & 33.26 & 39.49\\
        w/o CSMF & 769K & 31.71\% & 80.75G & 33.26 & 39.49\\
        \textbf{w/ CSMF} & 769K & 38.89\% & 79.30G & 33.26 & 39.50\\
        \hline
    \end{tabular}
    \caption{Ablation study on Cross-Shift Mask Fusion (CSMF). The bottom two lines represent PP-ATD-light performance without and with CSMF, respectively. \textit{Pure Pixels} represents the percentage of \textit{pure pixels} for PP-ATD-light, which is averaged on Manga109 x2 dataset. }
    \label{tab:ablation_shift}
\end{table}

\begin{table}
    \centering
    \fontsize{9}{11}\selectfont
    \begin{tabular}{|c|c|c|cc|}
        \hline
        Model & Params & FLOPs & Urban100 & Manga109\\
        \hline
        ATD-light & 769K & 87.15G & 33.26 & 39.49\\        
        w/o IPC & 769K & 79.30G & 33.20 & 39.43\\
        \textbf{w/ IPC} & 769K & 79.30G & 33.26 & 39.50\\        
        \hline
    \end{tabular}
    \caption{Ablation study on Information-Preserving Compensation (IPC). The bottom two lines represent PP-ATD-light performance without and with IPC, respectively.}
    \label{tab:ablation_compensation}
\end{table}

\subsection{Ablation Study}

% \subsubsection{Effects of HSV color space}

\subsubsection{Effects of Cross-Shift Mask Fusion (CSMF)}
As shown in table \ref{tab:ablation_shift}, our CSMF method increases the percentage of \textit{pure pixels} from 31.71\% to 38.89\%, saving 22.66\% more FLOPs compared to PP-ATD-light without CSMF (7.85G v.s. 6.4G). Notably, these computational savings are achieved without compromising model performance, as evidenced by the nearly identical PSNR scores on both Urban100 and Manga109 datasets. This compelling result validates the effectiveness of our CSMF strategy in optimizing computational resources while maintaining reconstruction quality.

\subsubsection{Effects of Information-Preserving Compensation}
As shown in table \ref{tab:ablation_compensation}, the Information-Preserving Compensation method effectively preserves the complete information volume and prevents the performance degradation, without introducing observable calculation burden. 

\subsubsection{More analysis} In the supplementary material, we provide more ablation studies and analysis.

\section{Conclusion}
In this work, we address key limitations of the existing implementation of Dynamic Token-Mixing Routing strategy for lightweight image super-resolution, such as poor adaptability, coarse-grained masking, and spatial inflexibility. By introducing Pure-Pass (PP), a pixel-level masking mechanism, we enable fine-grained, spatially flexible masking that exempts \textit{pure pixels} from costly computations while preserving adaptive flexibility. PP leverages fixed color center points to categorize pixels efficiently, ensuring minimal computational overhead. When integrated into the state-of-the-art ATD-light model, PP-ATD-light demonstrates superior SR performance, surpassing CAMixer-ATD-light in both reconstruction quality and parameter efficiency while maintaining comparable computational savings. Our approach advances the practical deployment of efficient SR models by balancing performance and complexity.

\bibliography{aaai2026}

\end{document}